%% file: Computer_Society_LaTeX_template/main.tex
\documentclass[journal]{IEEEtran}
\usepackage{amsmath,amsfonts}
\usepackage{array}
\usepackage[caption=false,font=normalsize,labelfont=sf,textfont=sf]{subfig}
\usepackage{textcomp}
\usepackage{stfloats}
\usepackage{url}
\usepackage{verbatim}
\usepackage{graphicx}
\usepackage{cite}
\usepackage{float} 
\usepackage{hyperref}
\usepackage{amsmath}
\usepackage{amssymb}
\usepackage{mathtools}
\usepackage{amsthm}

\input{math_commands.tex}

\usepackage{url}
\usepackage{booktabs}    
\usepackage{xcolor}      
\usepackage[table]{xcolor} 
\usepackage{graphicx}    
\usepackage{pifont} 
\usepackage{tabularx} 
\usepackage{array}
\usepackage{caption}   
\usepackage{siunitx}
\usepackage{multirow}   
\usepackage{colortbl}   
\usepackage{enumitem}
\usepackage{wrapfig}
\usepackage{subcaption}
\usepackage[export]{adjustbox} 
\usepackage{ragged2e}
\usepackage{makecell}
\usepackage{tabulary}   
\usepackage{tcolorbox}
\usepackage{seqsplit}

\tcbuselibrary{listings, breakable, skins} 
\definecolor{TiffanyBlue}{RGB}{129, 216, 208}

\definecolor{instructionbg}{RGB}{240, 245, 255} 
\newcommand{\codetext}[1]{\texttt{#1}}
\newcommand{\cmark}{\textcolor{black}{\checkmark}}
\newcommand{\xmark}{\textcolor{black}{\ding{55}}}

\usepackage[capitalize,noabbrev]{cleveref}
\theoremstyle{plain}

\theoremstyle{definition}

\theoremstyle{remark}

\usepackage[textsize=tiny]{todonotes}
\usepackage[ruled,vlined]{algorithm2e}
\usepackage{marvosym}

\begin{document}

\title{GraphThink: Graph-Enhanced LLM Thinking for Long-Horizon Embodied Task Planning}

%\author{IEEE Publication Technology,~\IEEEmembership{Staff,~IEEE,}}
        % <-this % stops a space
%\thanks{This paper was produced by the IEEE Publication Technology Group. They are in Piscataway, NJ.}% <-this % stops a space
%\thanks{Manuscript received April 19, 2021; revised August 16, 2021.}}
%\renewcommand*{\thanks}[1]{\protect\footnotemark}
%\protected\def\thanks{\protect\footnotemark}
\author{
Chen~Li$^{1,3}$,
Sijie~Cheng$^{4,6}$,
Yuelin~Zhang$^{1,3}$,
Junxi~Li$^5$,
Maozhi~Huang$^{1,3}$,
Yang~Liu$^4$,
and~Wenbing~Huang~\textsuperscript{\Letter}$^{1,2,3}$

\IEEEcompsocitemizethanks{
\IEEEcompsocthanksitem \textsuperscript{\Letter} denotes the corresponding author: hwenbing@ruc.edu.cn.%
% \thanks{$^1$ Gaoling School of Artificial Intelligence, Renmin University of China, Beijing 100872, China.}%
% \thanks{$^2$ School of Computer Science and Technology, Tsinghua University, Beijing 100084, China.}%
% \thanks{$^3$ Department of Electrical and Electronic Engineering, The Hong Kong Polytechnic University, Hong Kong SAR, China.}%
\IEEEcompsocthanksitem $^1$ Gaoling School of Artificial Intelligence, Renmin University of China, Beijing, China.
\IEEEcompsocthanksitem $^2$ Beijing Academy of Artificial Intelligence, Beijing, China.
\IEEEcompsocthanksitem $^3$ Beijing Key Laboratory of Research on Large Models and Intelligent Governance, Beijing, China.
\IEEEcompsocthanksitem $^4$ Department of Computer Science and Technology, Tsinghua University, Beijing 100084, China.
\IEEEcompsocthanksitem $^5$   Department of Electrical and Electronic Engineering, The Hong Kong Polytechnic University, Hong Kong SAR, China.
\IEEEcompsocthanksitem $^6$ RayNeo.AI, Shenzhen, China. 
}}

% The paper headers
%\markboth{Journal of \LaTeX\ Class Files,~Vol.~14, No.~8, August~2021}%
%{Shell \MakeLowercase{\textit{et al.}}: A Sample Article Using IEEEtran.cls for IEEE Journals}

%\IEEEpubid{0000--0000/00\$00.00~\copyright~2021 IEEE}
% Remember, if you use this you must call \IEEEpubidadjcol in the second
% column for its text to clear the IEEEpubid mark.

\maketitle

\begin{abstract}
Embodied agents using LLM-based planners often struggle with physical hallucinations, poor generalization to long-horizon tasks, and lack of environmental awareness. We propose \emph{GraphThink}, a novel framework that integrates a \emph{task graph} to provide structured knowledge for robust planning and a \emph{scene graph} to maintain environmental memory for event-driven replanning. Specifically, the task graph guides LLM thinking through contextual prompting and iterative refinement, effectively mitigating planning hallucinations. Furthermore, within the GRPO framework, the task graph offers delicate reward design to train the LLM planner, enhancing long-horizon planning capabilities and improving generalization. Finally, an event-driven replanning module, powered by the scene graph, enables closed-loop environment awareness and error correction. GraphThink achieves state-of-the-art performance on the ALFRED benchmark. In particular, our high-level planner surpasses leading API-based LLMs on both the validation set and held-out long-horizon tasks, underscoring its robust zero-shot and few-shot capabilities. Additional evaluations further demonstrate strong out-of-distribution generalization to novel tasks and environments.
\end{abstract}


\section{Introduction}

\IEEEPARstart{T}{here} has been a growing exploration of embodied agents designed to execute long-horizon everyday tasks given human instructions~\cite{zhang2024vlabench,kim2024online1,kim2024realfred2,cai2025cookbench}. In the field of Embodied AI, instruction following necessitates that agents perform three key operations: interpreting natural language, using egocentric visual observations, and executing physical action to navigate and interact with the environment.
A straightforward approach~\cite{pashevich2021episodic,suganuma2021look,ehsani2024spoc} involves training agents in an end-to-end supervised manner using large-scale datasets with annotated instructions and low-level expert action sequences. 
However, this paradigm is resource-intensive: it relies heavily on task-specific data and shows poor generalization to unseen scenarios.
In contrast, data-efficient hierarchical methods~\cite{min2021film,bhambri2023multi4,yang2024disco,kim2025flare} have emerged as a promising alternative: the high-level planner first decomposes instructions into subtasks, while the low-level executor subsequently accesses the skill library to translate these subtasks into executable actions in the environment. 

Recently, hierarchical methods have increasingly harnessed Large Language Models (LLMs) for high-level planning~\cite{ahn2022can,singh2023progprompt,rana2023sayplan,sun2025retrieval}, owing to their strong language understanding and reasoning capabilities. When initial planning fails, these models can engage in replanning by incorporating observed environmental objects as contextual prompts~\cite{song2023llm-planner,chen2025robogpt,kim2025flare}.
Nonetheless, long-horizon planning remains a major challenge for embodied agents, with three key limitations: 
(i) Despite their strong reasoning capabilities, general-purpose LLMs lack physical grounding, leading to instruction misinterpretation, planning hallucinations, and increased failure rates as task complexity grows.
(ii) Supervised fine-tuning on limited in-domain data results in poor generalization to unseen long-horizon tasks, except when employing prohibitively expensive annotations.
(iii) Many existing replanning strategies that primarily trigger corrections for low-level failures are prone to myopic decisions and inefficient retries, as they lack the environmental awareness to detect plans that are executable by low-level actions but misaligned with the instruction.

\begin{figure*}[t!]
    \centering
    \includegraphics[width=1\linewidth]{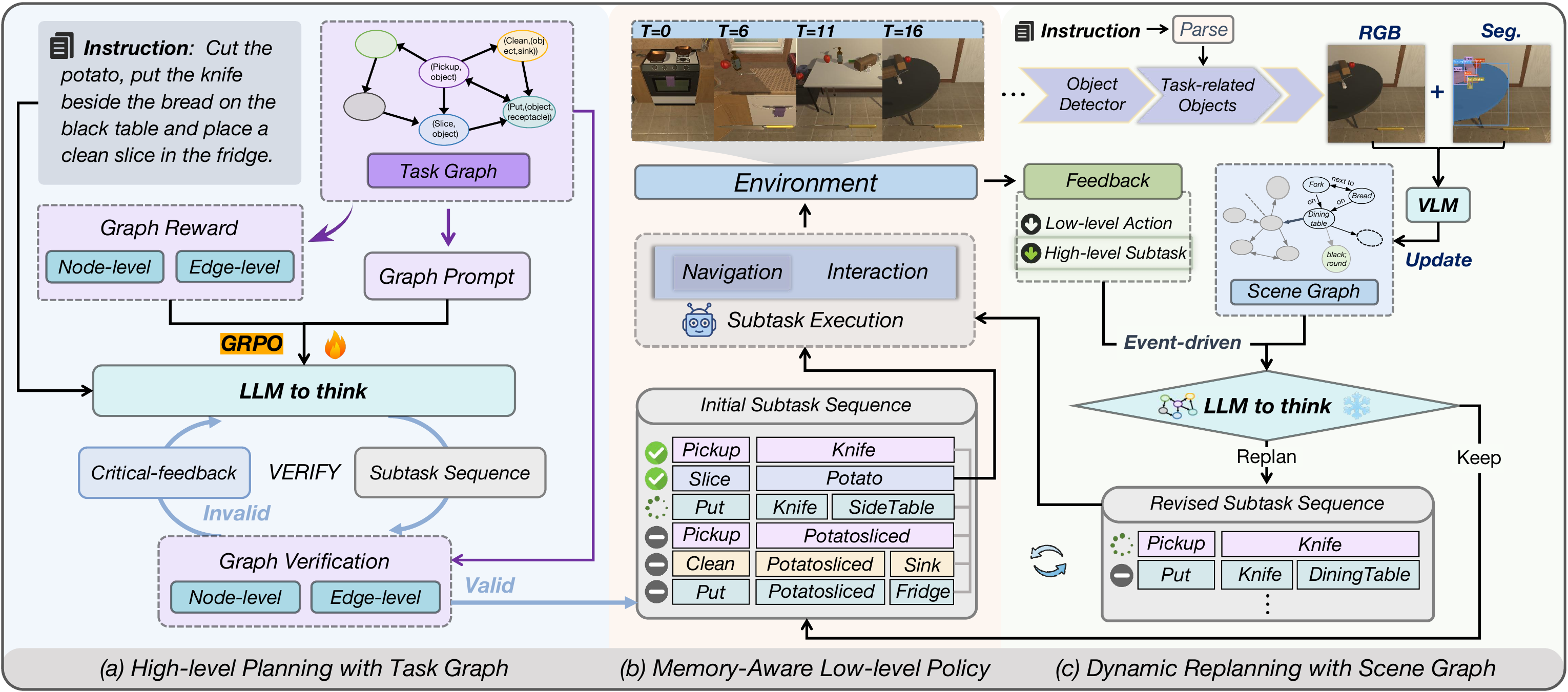}
    \caption{GraphThink consists of three core modules: (a) the high-level planner with task graph generates an initial plan, (b) the memory-aware low-level policy executes navigation and interaction actions for each subtask, and (c) dynamic replanning with scene graph adjusts plans during execution. The example here shows replanning triggered upon the completion of subtask ``put(knife, sidetable)", and the revised subtasks become ``pickup(knife)" and ``put(knife, dining table)". 
    %The subtask colors correspond to their respective task graph nodes.
    }
    \label{fig:framework}
    \vskip -0.2in
\end{figure*}

To address these challenges, we propose \textbf{\emph{GraphThink}}, \textcolor{black}{a dual-graph enhanced closed-loop planning system}, as illustrated in Fig.~\ref{fig:framework}.
This framework integrates two core structured representations: a \emph{task graph} to provide structured knowledge for robust planning, and a \emph{scene graph} to maintain environmental memory for event-driven replanning.
The \textbf{\emph{task graph}} plays three essential roles in high-level planning. First, it guides the LLM’s plan generation by explicitly incorporating the task graph into the prompt. Second, it serves as an external verifier to detect planning hallucinations and refine the subtask sequence iteratively. Finally, we adopt Group Relative Policy Optimization (GRPO)~\cite{shao2024deepseekmath} with task graph-based reward to enhance the reasoning capabilities of LLMs. Since embodied planning admits multiple valid solutions for a single goal, reward design becomes particularly challenging. The task graph addresses this issue by encapsulating diverse feasible paths, enabling reward signals that accommodate multiple valid solutions. With this delicate reward design, we are able to facilitate the alignment between the high-level action space and instructions, leading to stronger reasoning and generalization capabilities.

To enhance the LLM's environmental awareness and correct instruction misalignment within a closed-loop planning process, we design an event-driven replanning module powered by the scene graph. The dynamically updated \textbf{\textit{scene graph}} serves as \textcolor{black}{a task-centric memory module} to focus reasoning on task-relevant environmental cues. Unlike general scene graphs~\cite{gu2024conceptgraphs,takmaz2025search3d} that often include numerous irrelevant objects, our design keeps the graph size manageable, filtering out noise and enabling more efficient LLM reasoning. When replanning is triggered by low-level execution errors or proactive checks upon high-level subtask completion, the current scene graph memory and plan execution progress are provided to the thinking LLM to support efficient tracking and adaptation of the plan. To ensure the quality of replanning, all proposed revisions are constrained by the task graph to maintain feasibility. Together, these components form a closed-loop system that continuously aligns plan execution with user intent.

We evaluate GraphThink on ALFRED~\cite{shridhar2020alfred}, a challenging benchmark for vision-language navigation and interaction. Our hierarchical planning framework \textbf{ranks first on the official ALFRED leaderboard}. Particularly, we evaluate the performance of the high-level planner and \textbf{\textit{contribute a long-horizon dataset}} to better assess long-horizon generalization. The results show GraphThink outperforms various leading API-based LLMs on both the validation set and unseen long-horizon benchmark, demonstrating few-shot and zero-shot learning capabilities. \textcolor{black}{Further, GraphThink shows strong generalization to novel actions in AI2-Thor~\cite{kolve2017ai2} and effective cross-environment transfer to VirtualHome~\cite{puig2018virtualhome}.}

\section{Related Work}
\label{headings}

\noindent\textbf{Task Planning for Embodied Agents.}
Prior work~\cite{shridhar2020alfred,pashevich2021episodic,suganuma2021look,singh2021factorizing} train agents end-to-end to directly generate low-level actions given language instructions, but their performance in long-horizon tasks remains limited. 
Recently, hierarchical or modular planning~\cite{inoue2022prompter,shi2024opex,kim2025flare} have proven effective by decomposing tasks into subtasks to bridge the gap between natural instructions and executable actions.
In the early stage, template-based methods~\cite{min2021film,yang2024disco} are limited to predefined tasks and struggle to generalize.
To address this problem, LLMs are being explored as high-level planners, either through few-shot in-context prompting~\cite{song2023llm-planner,kim2025flare} or by supervised training on specific datasets~\cite{zhao2024epo, chen2025robogpt}.
\textcolor{black}{Some works~\cite{huang2022inner,song2023llm-planner,kim2024pre_replan3,kim2025flare}} further introduce replanning mechanisms to adjust actions by accepting environmental feedback, triggering local corrections to immediate errors \textcolor{black}{or predefined state differences. Hence}, we propose an event-driven dynamic replanning mechanism to enhance both plan feasibility and instruction alignment.

\noindent\textbf{Complex Reasoning with LLMs.}
To solve complex reasoning tasks, Chain-of-Thought (CoT) methods~\cite{wei2022chain,cheng2023unsupervised,obi2025safeplan} prompt LLMs to generate intermediate reasoning steps. However, as the number of steps increases, errors tend to accumulate. Self-correction methods~\cite{madaan2023selfrefine,guan2024amorfeedback} leverage feedback to refine incorrect reasoning and improve accuracy.
Moreover, Retrieval-Augmented Generation (RAG)~\cite{xu2024prag,wang2025instructrag} and knowledge graphs~\cite{wang2025double,zhu2025knowle-graph} methods enhance reasoning by integrating structured external knowledge to improve accuracy and reduce hallucinations.
To further enhance the performance through learning from interaction, recent work has shifted to supervised fine-tuning (SFT)~\cite{zhang2023yousft} and reinforcement learning (RL)~\cite{qian2025toolrl}. In our work, we further introduce a task graph-based RL framework that enhances the model's long-horizon planning capability through reward signals derived from the task graph.

\section{Methodology}

As illustrated in Fig.~\ref{fig:framework}, GraphThink mainly consists of three components: (a) High-level planning with task graph; (b) Memory-aware low-level action policy; (c) Dynamic replanning with scene graph memory. We provide the details of each component in this section.

\subsection{High-Level Planning with Task Graph}

To provide physical grounding for high-level planning, we introduce a task graph that models logical and executability constraints over subtask transitions. Rather than merely memorizing trajectories observed in the training data, we design an automatic task graph construction pipeline to instantiate feasible subtask-transition priors, which provide structured guidance for generating executable subtask sequences. The task graph is integrated across multiple stages of the high-level planning pipeline: it guides LLM planning through carefully designed prompts that incorporate the task graph, supports the design of reward functions in GRPO, and acts as an external verifier to provide feedback for reasoning.\\
\subsubsection{Task Graph Construction}\label{taskgraph_construct}
Subtasks in embodied planning often exhibit ordering dependencies due to physical preconditions (\textit{e.g.}, \texttt{Slice(object)} typically requires \texttt{Pickup(knife)} first), and we therefore represent feasible subtask transitions as a directed task graph $\mathcal{G}=(\mathcal{V},\mathcal{E})$, where nodes $\mathcal{V}=\{v_i\}_{i=1}^N$ denote subtasks and each edge $(v_i,v_j)\in\mathcal{E}$ indicates that $v_j$ can executably follow $v_i$. The \textbf{planning objective} is: given a natural language instruction $\tau$, high-level planning aims to generate a subtask sequence $S(\tau)=(s_1, s_2, \dots, s_T)$ that guides the agent to complete the task as instructed. Here, the subtask sequence should form a feasible path $\Pi = (v_{1}, v_{2}, \dots, v_{T})$ in $\mathcal{G}$, and each object in $s_i$ belongs to the environmental object set $\mathcal{O}_{\text{env}}$, which comprises all visible objects recognized by a pretrained object detection module.

\textbf{Subtasks as Nodes.}
Node \(v_i\) representing a subtask is denoted as \(A_i(o_i)\) or \(A_i(o_i,o_j)\), where \(A_i\) denotes a high-level action and \(o_i,o_j\) refer to relevant objects. Each subtask is grounded in the available low-level skill set, thereby mapping the LLM's open-ended text generation to robot-executable actions. 
Following prior work~\cite{min2021film,yang2024disco}, our evaluation on the ALFRED benchmark uses 12 subtask nodes as an appropriate granularity for high-level planning to executable low-level policies. 
Given the vast combinatorial space of subtask--object pairs, we employ meta-classes 
\(\mathcal{C}\coloneqq\{\texttt{obj},\texttt{rec},\texttt{mov},\texttt{spec}\}\) 
to abstract away specific object instances, where \texttt{obj} denotes pickupable items, \texttt{rec} designates fixed containers, \texttt{mov} indicates portable containers, and \texttt{spec} denotes task-specific operation targets required by certain subtasks. 
For example, \texttt{spec} can be instantiated as Lamp for \texttt{ToggleObject}, Fridge for \texttt{CoolObject}, Microwave for \texttt{HeatObject}, and SinkBasin for \texttt{CleanObject}. 
This shifts the planner’s focus to functional matching between subtasks and objects, thereby facilitating generalization across environments among different objects with similar functionalities. During plan generation, meta-classes are instantiated as concrete object names from the environmental object set \(\mathcal{O}_{\text{env}}\).
During execution, each high-level subtask is expanded into a serialized low-level action policy and dynamically instantiated through environmental perception. For example, \texttt{Heat(object, microwave)} is not a single primitive action but a policy involving navigation, object placement, microwave operation, and object retrieval.

\textbf{Subtask Transition as Edges.} A task graph can be constructed by traversing training trajectories. However, trajectory-based construction only captures demonstrations observed in the dataset and may therefore miss feasible yet unobserved subtask transition paths.
To build a more general task graph, we construct edges through LLM-assisted transition compatibility analysis, which considers both semantic compatibility and resource-state feasibility between subtasks. Specifically, for each subtask \(v_i\), the LLM receives its serialized low-level policy and abstracts it into a structured transition interface \(\langle P_i, E_i\rangle\), where \(P_i\) denotes the preconditions required to start the subtask and \(E_i\) denotes the state or resource effects after completing it. We add a directed edge \(v_p \to v_q\) when the effects \(E_p\) of the preceding subtask satisfy the preconditions \(P_q\) of the subsequent subtask, and the transition between the two subtasks is semantically valid. For example, after \texttt{PickupObject(obj)}, the resulting effect includes \(\texttt{holding(obj)}\) and hand-resource occupation, which satisfies the precondition of \texttt{PutObject(obj, rec)} that the object should be held. This mechanism supports generalization to novel tasks without requiring trajectory demonstrations. 

\textbf{Scalability of Task Graph.} 
The task graph is designed to be easily extensible. 
When generalizing to new tasks or environments (Sec.~\ref{novel_action_env}), new subtask nodes can be introduced according to specific task requirements as the low-level policy set evolves. 
Importantly, their incident edges can be generated through the same transition compatibility analysis, enabling task-graph extension without requiring new trajectory demonstrations. This design allows practitioners to choose an appropriate subtask granularity according to the target environment and task requirements. Appendix A further verifies the reliability of this construction strategy and its robustness to noisy edge constraints.\\
\subsubsection{Prompt with Task Graph}
To mitigate planning hallucinations in LLM-based task planning, we propose a novel prompting strategy incorporating robotic domain knowledge via a task graph. This strategy directs the agent to generate subtask sequences that follow the graph topology, where each subsequent subtask is guided to follow a neighboring node of previous subtask. Specifically, the nodes are encoded as action-target pairs assigned unique labels (\emph{e.g.}, `[A] PickupObject(object)'), while the edges are converted into symbolic connection relationships (\emph{e.g.}, `A → B'). Static prompt components (e.g., agent roles, planning rules, output format) are also incorporated to facilitate in-context learning. The complete prompt template is shown in Appendix G.\\
\subsubsection{GRPO with Task Graph}
Recent work~\cite{qian2025toolrl,vojnovic2025alignment} demonstrates that rule-based rewards combined with RL significantly enhance reasoning. For embodied task planning, reward design is challenging because a single instruction may admit multiple feasible solutions. 
We address this by combining graph-based structural rewards with instruction-following rewards in GRPO. 
Specifically, \textit{\textbf{task graph rewards encourage exploration over feasible task-graph paths}}, thereby avoiding dependence on a single annotated trajectory, while \textit{\textbf{instruction rewards serve as a semantic constraint}} that selects instruction-aligned plans among feasible candidates. 
This complementary design improves generalization while preventing reward hacking, such as generating plans that satisfy task-graph constraints but deviate from the instruction semantics. We next describe the detailed reward design used in GRPO.

\textbf{Format Reward.} 
Prior research~\cite{guo2025deepseek,xie2025logic} has shown that the format reward  $R_{\text{fmt}}$ effectively constrains the model's output to match the expected structure. We customized the format reward to suit the output structure of high-level task planning. The format reward is computed as:
 \begin{equation}
 R_{\text{fmt}} = 
 \begin{cases}
 1.0, & \text{if \codetext{XML} and \codetext{JSON} valid}, \\
 0.5, & \text{if \codetext{XML} valid but \codetext{JSON} invalid}, \\
 0.0, & \text{otherwise}.
 \end{cases}
 \end{equation}
 $R_{\text{fmt}}=0.5$, if the model's response follows \codetext{XML\_format}, \emph{i.e.}, \textit{\textless think\textgreater\ \ldots\ \textless /think\textgreater \textless answer\textgreater\ \ldots\ \textless /answer\textgreater}; $R_{\text{fmt}}=1$, only when the content inside \textit{\textless answer\textgreater\ \ldots\ \textless /answer\textgreater} is a \codetext{JSON\_object} including the specified key, \emph{e.g.}, ``high\_level".
   
\textbf{Node-Level Reward.}
To encourage each generated subtask $s_i$ to be valid, we compute a node-level reward via dynamically weighted object-validity scores across four meta-classes in $\mathcal{C}$: 
\begin{equation}
\textstyle R_{\text{node}} = \sum_{c \in \mathcal{C}} w_c \cdot R_{\text{node}}^c,
\end{equation}
where the weight $w_c$ is dynamically adjusted based on whether the corresponding meta class $c$ appears in the plan. Weights of unused meta classes are set to zero, and the remaining weights are renormalized to ensure that all weights sum to 1.
For each meta class, the reward \(R_{\text{node}}^c\) is defined as the validity rate of objects belonging to meta class \(c\) in the plan, where an object is considered valid when its name exists in the corresponding environmental list $\mathcal{O}_{\rm env}^{c} $, then:
\begin{equation}
\textstyle R_{\text{node}}^c = \frac{1}{|\mathcal{O}_{S}^c|} \sum_{o \in \mathcal{O}_{S}^c} \mathbb{I}\left(o \in \mathcal{O}_{\text{env}}^c\right),
\end{equation}
where $\mathcal{O}_{S}^c$ denotes the set of all objects in plan $S(\tau)$ that correspond to subtasks of class $c$, and $\mathbb{I}(\cdot)$ is an indicator function that returns 1 if the condition holds and 0 otherwise. 

\textbf{Edge-Level Reward.}
To encourage subtask transitions conforming to the edges $\mathcal{E}$ in the task graph $\mathcal{G}$, this reward traverses and checks all adjacent subtask pairs $(s_{t}, s_{t+1})$ in the generated plan. The reward is defined as follows:
\begin{equation}
\textstyle R_{\text{edge}} = \prod_{\substack{1 \leq t \leq T-1}} \mathbb{I}(s_{t}, s_{t+1} \in \mathcal{V}) \cdot  \mathbb{I}((s_{t}, s_{t+1}) \in \mathcal{E}),
\label{edge_reward}
\end{equation}
where the reward equals 1 if both subtasks \((s_t,s_{t+1})\) are nodes in \(\mathcal{V}\) and the transition between them satisfies the feasible-transition constraint of \(\mathcal{G}\), and 0 otherwise.

\textbf{Instruction Following Reward.}
The reward for instruction following $R_{\text{inst}}$ encourages the model to generate plans that satisfy user instructions. Unlike tasks with unique solutions (\emph{e.g.}, arithmetic or classification), task planning is inherently multi-solution, making exact-match evaluation insufficient. To address this, we employ a two-stage reward based on the Ground-Truth (GT) plan $S^*$. First, we extract critical subtasks $\mathcal{S}^*_{\text{key}}$ from $S^*$ that are essential for completing the instruction. The generated plan $S(\tau)$ receives a reward of 0.5 if it covers all critical subtasks, and a full reward of 1.0 only if it exactly matches the GT plan $S^*$. Otherwise, the reward is 0. This two-stage design ensures that the model is first driven to cover indispensable steps and then incentivized to produce fully precise plans.

\textbf{Overall Reward Function.}
The overall reward function integrates rewards to guide the model in generating executable subtask sequences that meet instruction goals:
\begin{equation}
\textstyle R_{\text{total}} = R_{\text{fmt}} + R_{\text{node}} + R_{\text{edge}} + R_{\text{inst}}.\\
\end{equation}

\subsubsection{Verification with Task Graph}
Inspired by \cite{gou2023critic}, we treat the task graph as an external validation tool and leverage its feedback to refine the responses of the LLM. Since task-graph constraints have already been learned through graph-based rewards during GRPO, verification mainly serves as a lightweight safeguard at inference to correct occasional illegal transitions and reduce residual planning hallucinations. The verification function is composed of two parts:

\textbf{Node-Level Legitimacy Verification}. To determine whether each object in the initial subtask sequence is valid, we perform a node-level legitimacy verification:
$ \mathbb{V}_{\text{node}} = \prod_{c \in \mathcal{C},o \in \mathcal{O}_{S}^c} \mathbb{I}(o \in \mathcal{O}_{\text{env}}^c)$.

\textbf{Edge-Level Legitimacy Verification.} 
To verify that all action transitions satisfy the edge constraints of the task graph, the edge-level legitimacy check is defined analogously to the edge-level reward $R_{\text{edge}}$ (see Eq. (\ref{edge_reward})):
$\mathbb{V}_{\text{edge}} = R_{\text{edge}}$.

For any invalid subtask or transition, the corresponding error message is fed back to guide LLM correction. This \textit{Verify} → \textit{Feedback} → \textit{Correct} cycle iterates until the plan passes validation or reaches the maximum iteration limit (\emph{i.e.} 3).

\subsection{Memory-Aware Low-Level Action Policy}
\begin{figure}[t!]
    \centering
    \includegraphics[width=0.98\linewidth]{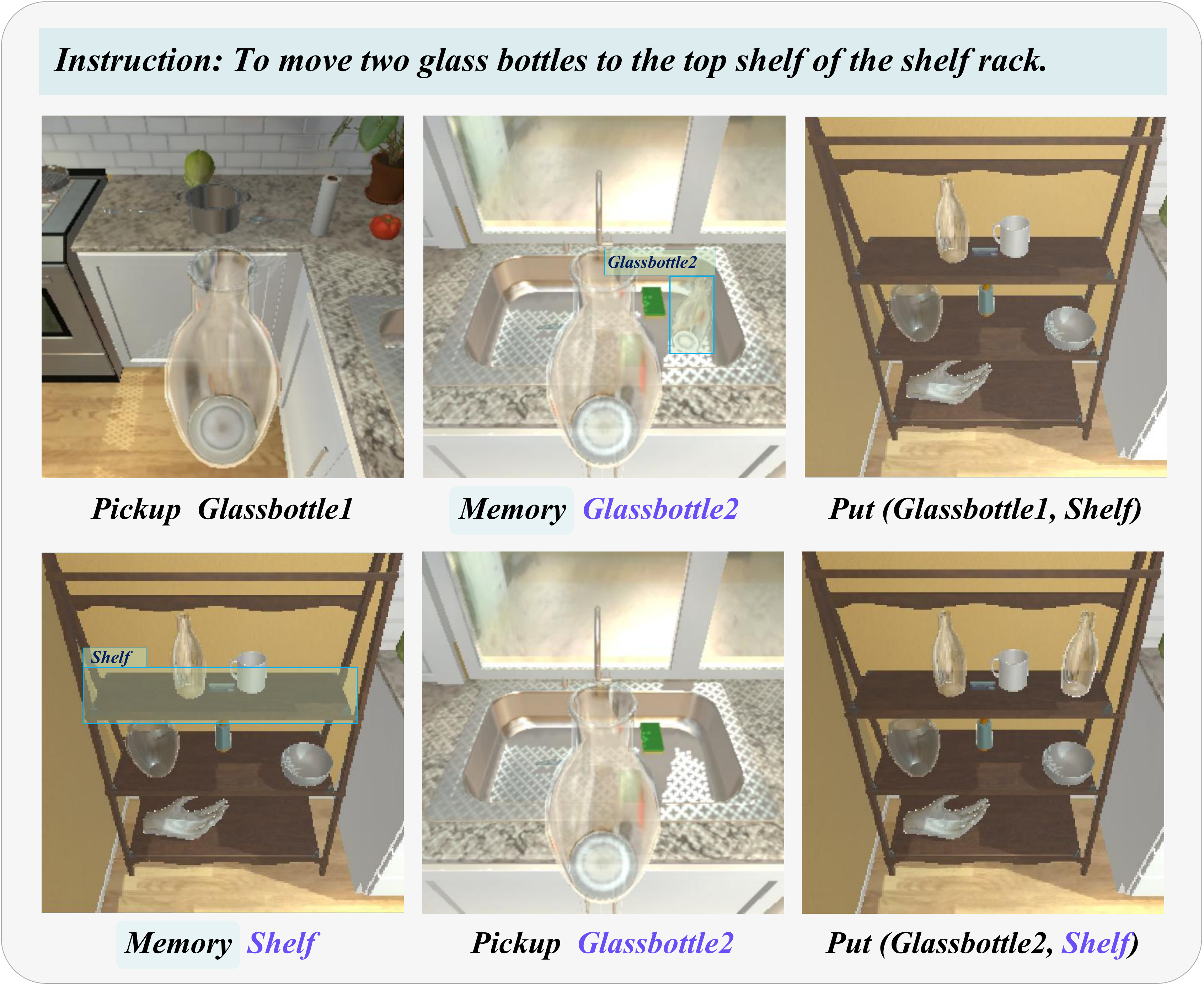}
    \caption{Qualitative example illustrating the benefits of the memory-aware low-level action policy.}
    \label{fig_envmem}
\end{figure}
The low-level policy translates high-level subtasks into executable primitive actions, dynamically grounded through environmental perception. To enhance navigation efficiency and prevent redundant interactions, we develop a \textit{memory-aware low-level action policy}. \textcolor{black}{While prior works (\textit {e.g}.,~\cite{kim2023context}) have explored memory-enhanced low-level policy, our approach advances in proactive caching and instance-level discrimination.} \textcolor{black}{Unlike methods that primarily record past interactions, our agent proactively detects and caches information for objects that may require future interaction during execution.} When the agent encounters other task-related objects on the way to the primary target object, it logs its current location as a candidate waypoint and caches the corresponding segmentation mask, in addition to saving the position and mask of the target object. Thus, subsequent subtasks can directly utilize cached positions instead of re-exploring, improving navigation consistency and efficiency.
For distinguishing multiple instances of the same type, individual interaction records are maintained to ensure precise identification and consistent placement. As shown in Fig.~\ref{fig_envmem}, this memory mechanism allows detecting and recording secondary `Glassbottle2' during the execution of `\texttt{Pickup(}Glassbottle1\texttt{)}'. Subsequently, when executing `\texttt{Put(}Glassbottle2, Shelf\texttt{)}', the agent prioritizes the shelf position retrieved from memory instead of re-exploring, improving navigation efficiency and placement accuracy.

\subsection{Dynamic Replanning with Scene Graph}
Although high-level planning with the task graph yields high-quality initial plans, it is not well grounded in dynamic environments and cannot readily adapt to newly emerged failures. To address this, we design an event-driven dynamic replanning module grounded by a dynamic scene graph, which invokes a replanning reasoning process upon either low-level execution errors or high-level subtask completion to determine whether the current plan should be kept or revised.\\
\subsubsection{Task-Aware Scene Graph Memory}
\label{sec:scene_graph_memory}
To support closed-loop reasoning and event-driven replanning, GraphThink maintains a dynamic scene graph 
\(\mathcal{G}_{\rm scene}\) as a task-aware environmental memory. Unlike complete scene representations~\cite{gu2024conceptgraphs,takmaz2025search3d}, \(\mathcal{G}_\text{scene}\) is a compact, dynamically updated set of semantic triples \((o_i, r_{ij}, o_j)\) that describe relations $r_{ij} \in \mathcal{R}$ (\emph{e.g.}, ``on'', ``next to'') between task-relevant objects, with key attributes (\emph{e.g.}, color and shape) as node attributes. This task-focused representation reduces irrelevant visual noise and keeps the graph scale manageable 
(typically fewer than 20 nodes in ALFRED), while retaining the environmental evidence required for replanning.
The construction of \(\mathcal{G}_{\rm scene}\) proceeds online during execution and does not require a pre-built scene graph:
(i) \textbf{Key Object Extraction.}
A language model parses the instruction $\tau$ to obtain \(\mathcal{O}_\text{key} \subseteq \mathcal{O}_\text{env}\).
(ii) \textbf{Viewpoint Capture.}
During execution, record the agent’s egocentric views that contain any object in $\mathcal{O}_{\rm key}$. To avoid unreliable relation captures from extreme viewpoints, we only perform relation extraction when the target object lies in the agent's forward-facing view, i.e., the angular deviation between the agent's heading and the direction to the object center is within \(45^\circ\). 
(iii) \textbf{Candidate Triple Generation.} The captured RGB image and the corresponding segmentation map are jointly provided to the VLM. The segmentation map helps align VLM-generated labels with environment object instances and reduces ambiguity caused by synonymous or visually similar objects. The VLM then produces candidate relation triples and object attributes. 
(iv) \textbf{Scene Graph Update.} Candidate entries are filtered and incrementally merged into \(\mathcal{G}_{\rm scene}\) as execution proceeds. 
During execution, interaction-induced changes also update the graph by replacing outdated relations, enabling the memory to adapt to object state changes and post-interaction environmental updates.

To reduce the impact of VLM hallucinations and perception errors, GraphThink adopts preventive filtering and conflict-aware correction. 
Before insertion, candidate entries are filtered by task relevance and rule-based physical-plausibility constraints; unreliable triples, such as 
\((\texttt{SideTable}, \texttt{on}, \texttt{CounterTop})\), or task-irrelevant entries are pruned. 
When a new observation conflicts with existing relations, GraphThink does not keep mutually inconsistent entries simultaneously. 
Instead, the agent re-observes the scene from another viewpoint and asks the VLM to reassess the conflicting entries. If some errors still enter the scene graph, they usually appear as execution failures or later contradictory observations, which trigger our event-driven replanning module and subsequent scene graph updates. This prevents corrupted memory from silently accumulating over long horizons. 

In this way, \(\mathcal{G}_{\text{scene}}\) serves not as a static map, but as a dynamically corrected memory that provides compact and reliable grounding for downstream replanning.\\
\subsubsection{Event-driven Replanning}
\begin{figure}[t!]
% \vskip -0.2in
\centering
%\vspace{-0.2cm}
\includegraphics[width=0.48\textwidth]{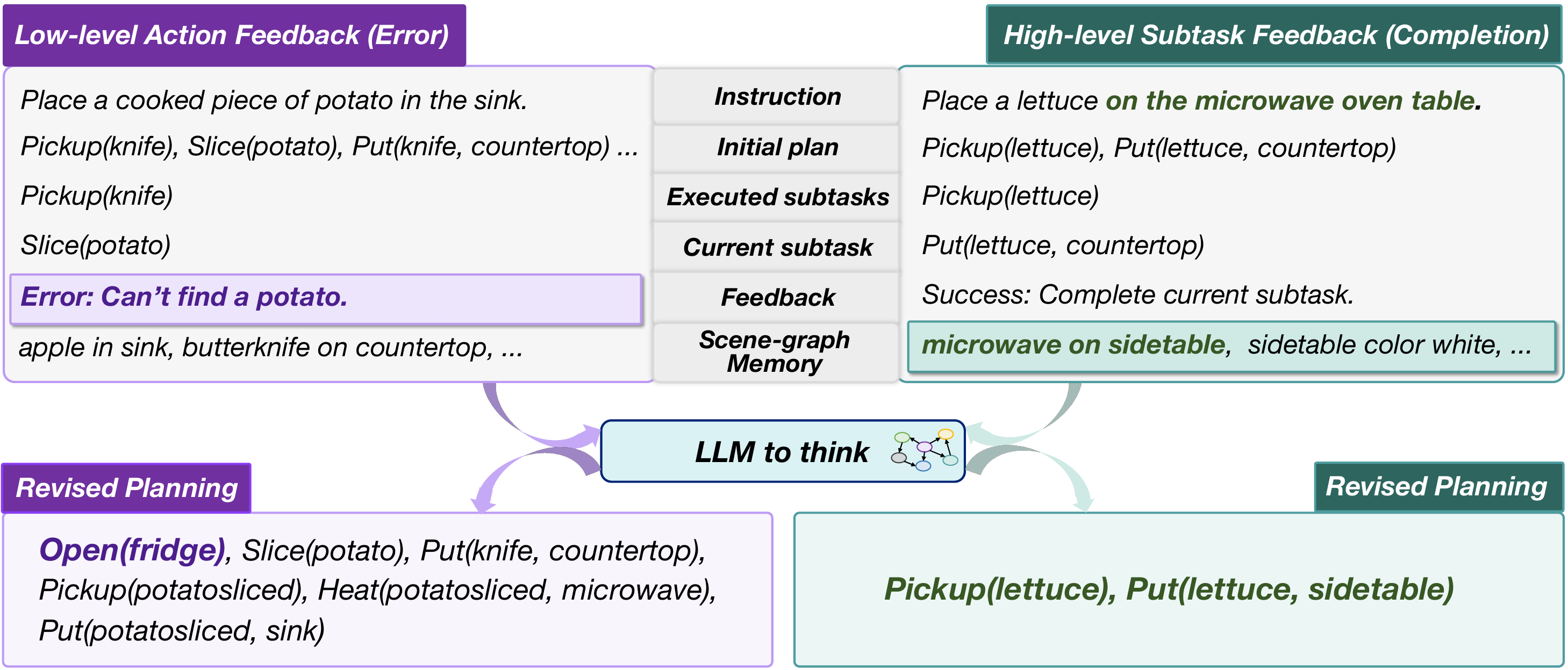} % Reduce the figure size so that it is slightly narrower than the column. Don't use precise values for figure width.This setup will avoid overfull boxes.
%\vskip -0.1in
\caption{The two examples illustrate two types of event-driven dynamic replanning.}
\label{fig_dynamic}
%\vskip -0.2in
\end{figure}

We aim for the agent to interleave execution with reasoning, continuously evaluating plan feasibility against environmental feedback and making necessary adjustments. 
Unlike recent planning methods that primarily revise plans after execution failures~\cite{huang2025one,kim2025flare}, GraphThink uses an LLM-based replanning module driven by two complementary events: low-level action errors and high-level subtask completion (Fig.~\ref{fig_dynamic}). 
The former addresses actionable execution failures, while the latter proactively checks semantically valid but goal-misaligned plans at subtask checkpoints.
Within this module, the task graph constrains static transition feasibility over executable subtasks, and the scene graph provides task-relevant environmental context from dynamic observations, thereby reducing unconstrained LLM free-generation during replanning. 
When activated, the model assesses whether the plan should be adjusted based on the original plan, current subtask, feedback, and scene graph. Execution continues if no adjustment is needed, or switches to the revised plan otherwise. 

\begin{figure}[t!]
    \centering
    \includegraphics[width=\linewidth]{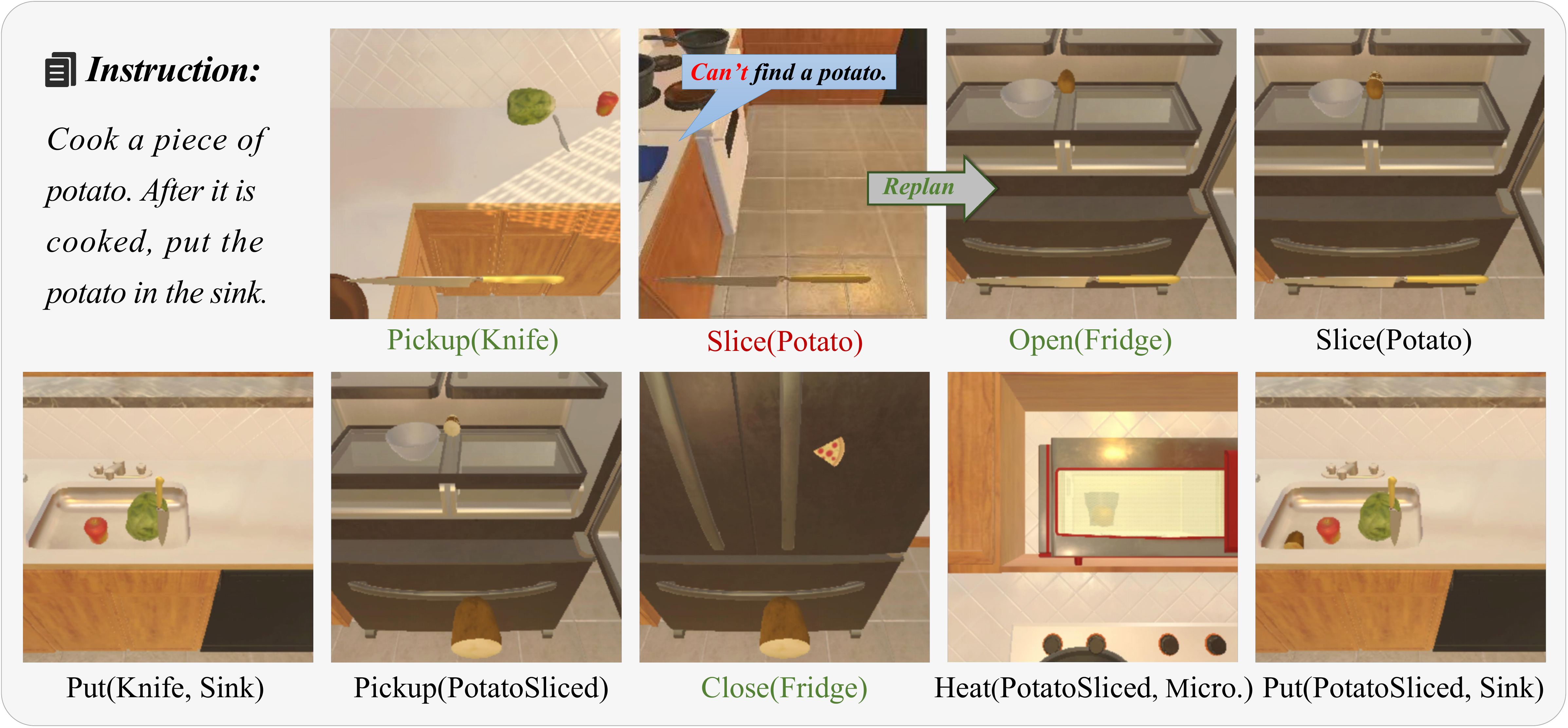}
    \caption{An example of error-triggered replanning. When the agent fails to locate a potato during environmental exploration while executing slice(potato), GraphThink revises the plan and then attempts to open the refrigerator to search for one.}
    \label{fig:replan_example_vla}
\end{figure}

\textbf{Low-Level Action Error.} 
This trigger handles concrete execution failures, detected via feedback signals from the environment (\emph{e.g.}, interaction errors) or from the navigation policy (\emph{e.g.}, exhaustive search failure). When such an error occurs, the reasoning LLM integrates the current scene observations with commonsense knowledge to revise the plan. For the instance in Fig.~\ref{fig_dynamic}, if a potato cannot be found for slicing, the LLM might suggest searching inside closed containers like a refrigerator or microwave. Notably, replanning is selectively invoked rather than triggered by every execution error.
Common recoverable failures, such as navigation collisions, are first handled by the low-level policy, and only unresolved cases are escalated to the replanning module. Detailed execution of this case is shown in Fig.~\ref{fig:replan_example_vla}. 

\textbf{High-Level Subtask Completion.} This trigger detects executable but goal-misaligned plans through LLM-based semantic assessment at subtask checkpoints.
The LLM assesses the current plan against the continuously updated scene graph memory and the plan execution progress to verify alignment with the final objective. For example, for the goal ``place the lettuce on the microwave oven table," if the scene graph indicates the microwave is on a side table but the plan targets a countertop, the LLM identifies this mismatch and triggers a correction. This allows proactive rectification of high-level planning errors undetectable by low-level error feedback, using subtask completion as checkpoints.

\section{Experiments}
In this section, we first validate GraphThink’s hierarchical framework for vision-language navigation and interaction, and further assess the performance of its high-level planner.

\begin{table}[t]
\centering
\caption{Comparison with SOTA methods on ALFRED test set. Baseline results are from the official leaderboard or papers. The ``Step-by-step Inst.'' column denotes whether step-by-step instructions are used in high-level planning.}
\label{comparsion}
\small
\setlength{\tabcolsep}{0pt}
\renewcommand{\arraystretch}{1.12}
\begin{tabular*}{\columnwidth}{@{\extracolsep{\fill}}l c c c c c@{}}
\toprule
\multirow{2}{*}{\textbf{Method}} 
& \multirow{2}{*}{\makecell{\textbf{Step-by-step}\\\textbf{Inst.}}} 
& \multicolumn{2}{c}{\textbf{Tests Seen}} 
& \multicolumn{2}{c}{\textbf{Tests Unseen}} \\
\cmidrule(lr){3-4}\cmidrule(l){5-6}
& & \textbf{SR $\uparrow$} & \textbf{GC $\uparrow$} 
  & \textbf{SR $\uparrow$} & \textbf{GC $\uparrow$} \\
\midrule
HLSM~\cite{blukis2022persistent}           & \cmark & 29.94 & 41.21 & 20.27 & 30.31 \\
LLM-planner~\cite{song2023llm-planner}     & \cmark & 18.20 & 26.77 & 16.42 & 23.37 \\
CAPEAM~\cite{kim2023context}               & \cmark & 52.58 & 60.98 & 50.36 & 61.40 \\
DISCO~\cite{yang2024disco}                 & \cmark & 59.59 & 66.06 & 56.55 & 66.87 \\
Flare~\cite{kim2025flare}                  & \cmark & 40.05 & 48.84 & 40.88 & 51.72 \\
\midrule
Prompter~\cite{inoue2022prompter}          & \xmark & 47.95 & 56.98 & 41.53 & 53.69 \\
DISCO~\cite{yang2024disco}                 & \xmark & 58.05 & 64.96 & 54.77 & 65.56 \\
OPEx~\cite{shi2024opex}                    & \xmark & 43.51 & 54.27 & 41.27 & 53.82 \\
EPO~\cite{zhao2024epo}                     & \xmark & 64.79 & 72.30 & 62.35 & 67.52 \\
RoboGPT~\cite{chen2025robogpt}             & \xmark & 59.92 & 67.83 & 62.00 & 72.09 \\
\midrule
\textbf{GraphThink} (ours)                 & \xmark & \textbf{67.71} & \textbf{73.96} & \textbf{68.52} & \textbf{75.76} \\
\bottomrule
\end{tabular*}
%\vskip -0.1in
\end{table}

\subsection{Comparison with SOTA Methods on ALFRED}
\noindent\textbf{Benchmark and Metrics.}
We conduct experiments on ALFRED~\cite{shridhar2020alfred}, a challenging benchmark for robotics instruction following. The language instruction $L=({L_\text{high}, L_\text{low}})$ contains both high-level goals and step-by-step guidance.
%The benchmark comprises 120 indoor scenes with diverse interactive objects and receptacles, supporting complex tasks that involve navigation, object manipulation, and multi-step reasoning.
ALFRED is partitioned into `train', `validation' and `test' sets.
Both `validation' and `test' are further divided into seen and unseen splits, where the unseen partitions consist of scenes absent from the training set. The dataset includes 7 task types, 58 target object classes, and 26 receptacle classes distributed across 120 indoor scenes. Objects within the same class often exhibit various visual appearances (\emph{e.g.}, there are 30 varieties of apples), and the indoor scenes cover kitchens, bathrooms, bedrooms, and living rooms. According to the official statistics, the training set consists of 21023 examples, the validation seen set contains 820 examples, the validation unseen set contains 821 examples, the test seen set has 1533 examples, and the test unseen set has 1529 examples. We adopt two evaluation metrics.
The primary metric is Success Rate (SR), which measures the percentage of fully completed tasks.
Additionally, Goal-Condition (GC) success rate evaluates the percentage of satisfied goal conditions. 
%Test set results require leaderboard submission.

\noindent\textbf{Performance.} To demonstrate the effectiveness of our approach, we compare GraphThink to competitive works in the test set reported on ALFRED public leaderboard. Following~\cite{kim2025flare,chen2025robogpt}, we report baselines using 1) only the goal instruction, and 2) both the goal instruction and step-by-step instructions. As shown in Table~\ref{comparsion}, our approach significantly outperforms prior works by 6.17 percentage points on unseen tasks, while achieving SOTA performance across all metrics in both unseen and seen environments (reaching 67.71\% and 68.52\%, respectively), demonstrating the effectiveness of our approach. Moreover, using only the goal instruction, GraphThink outperforms prior methods under both settings, i.e., with and without step-by-step instructions, highlighting the superiority of our high-level planning module. It is worth emphasizing that GraphThink does not rely on metadata for environmental grounding, indicating its stronger generalization capability to real-world scenarios.

\subsection{Ablations of GraphThink}

\begin{table}[t]
  \centering
  \small
  \caption{Ablation studies on three components of GraphThink's hierarchical framework.}
  \label{tab:ablation_GraphThink}
  \setlength{\tabcolsep}{10pt}
  %\vspace{0.5cm}
  %\vskip -0.1in
  \begin{tabular}{lccccc}
    \toprule
    \textbf{Component} & \multicolumn{2}{c}{\textbf{Valid Seen}} & \multicolumn{2}{c}{\textbf{Valid Unseen}} \\ 
    \cmidrule(lr){2-3} \cmidrule(lr){4-5}
     & \textbf{SR $\uparrow$} & \textbf{GC $\uparrow$} & \textbf{SR $\uparrow$} & \textbf{GC $\uparrow$} \\ 
    \midrule
    replaced planner & 40.70 & 46.19 & 41.59 & 47.82\\
    w/o replan             & 60.18 & 68.39 & 59.82 & 68.10 \\
    w/o replan\_low        & 64.31 & 72.13 & 64.10 & 71.58 \\
    w/o replan\_high       & 63.58 & 71.65 & 64.25 & 72.10 \\
    w/o memory             & 67.26 & 74.39 & 66.59 & 74.13 \\
    \midrule
    GraphThink              & \textbf{68.36} & \textbf{75.32} & \textbf{68.72} & \textbf{76.01} \\
    \bottomrule
  \end{tabular}
  \vskip -0.1in
\end{table}

We conduct an ablation study of GraphThink's components, with experimental results on the ALFRED validation set shown in Table~\ref{tab:ablation_GraphThink}.
(i) Replacing our task graph-based planner  with a RAG-enhanced Qwen2.5-7B-Instruct planner (Row 1) leads to significant performance drops of 27.66\% and 27.13\% on seen and unseen splits, respectively, underscoring the importance of high-quality initial planning.
(ii) Removing the replanning module results in open-loop execution that cannot adapt to unexpected errors or environmental feedback, reducing success rates by 8.18\% and 8.90\%. We then separately ablate the two types of event-driven replanning. Without low-level execution error driver (Row 3), the agent fails to handle execution errors (e.g., wandering due to objects in closed containers or detection failures), reducing SR by 4.05\% and 4.62\%. Without high-level subtask completion driver (Row 4), the agent cannot verify and correct plans against environmental information, causing SR drops of 4.78\% and 4.47\%.
(iii) Disabling the object state and location memory in the low-level policy (Row 5) degrades performance by impairing object localization and introducing interaction errors. In tasks requiring placement of two identical objects into the same container, the agent may move one object repeatedly or misplace them in different locations, resulting in task failures.
These results demonstrate that GraphThink ensures robust execution through: task graph-based planning for initial plans, dynamic replanning for maintaining executability and instruction alignment, and memory-aware navigation for reliable interaction.
\subsection{Evaluations of High-Level Planning}
This subsection evaluates GraphThink's high-level planner by comparing it with strong LLM-based planners and conducting extensive ablation studies on its key components and training design.\\
\subsubsection{Comparison with SOTA planners}\label{high-level-results}
\begin{figure*}
    \centering
    %\vspace{-0.2cm}
    \includegraphics[width=1\linewidth]{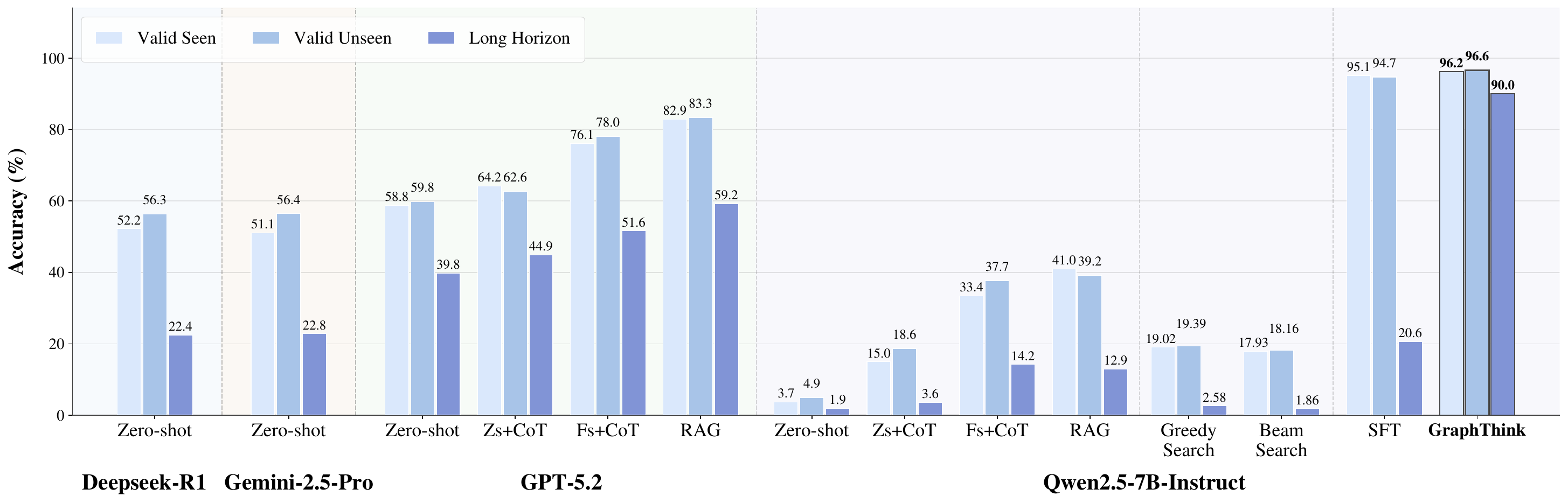}
    \vskip -0.1in
    \caption{Success rates for high-level planning on ALFRED validation set and long-horizon tasks. For brevity, we adopt the abbreviations Zs for Zero-shot and Fs for Few-shot.}
    \label{fig:plandata}
    %\vskip -0.2in
\end{figure*}
Since ALFRED has limited task diversity, we construct a long-horizon dataset of 1396 samples by extending original 7 short task types with 17 complex types to better evaluate unseen long-horizon planning (details in Appendix B). To evaluate GraphThink's high-level planner, we compare GraphThink against multiple strong LLMs (\emph{i.e.}, Deepseek-R1, Gemini-2.5-Pro, and GPT-5.2) under diverse inference settings (\emph{i.e.}, zero-shot, CoT, few-shot, RAG). For fair comparison, we conduct controlled evaluations on the same Qwen2.5-7B-Instruct backbone, comparing GraphThink with multiple inference strategies, SFT, and task-graph search baselines (Appendix C). 
For high-level planning, success rate is the primary metric. As a complementary evaluation, Section~\ref{Varying_Horizons} employs multiple fine-grained metrics to validate GraphThink’s robustness, efficiency, and reasoning quality across increasingly complex tasks. Since multiple feasible paths often exist in task planning and certain GT annotations in datasets may not accurately match the instructions, relying solely on the single GT in the validation set is insufficient. Therefore, we incorporate the task graph as a verification tool and propose a more comprehensive validation method. The procedure begins with graph-constrained verification and GT matching. If both checks pass, the plan is successful. If graph verification fails, the plan is marked as failed. When the plan satisfies graph constraints but diverges from GT, it undergoes further validation using an LLM.

\textbf{How does GraphThink compare to state-of-the-art LLMs?} As shown in Fig.~\ref{fig:plandata}, GraphThink achieves superior performance across all settings, with 96.22\% on valid seen, 96.56\% on valid unseen and 90.04\% on long horizon tasks. Although prompt-driven variants such as few-shot CoT and RAG yield substantial improvements upon stronger LLMs like GPT-5.2, GraphThink still surpasses all such methods by a significant margin.

\textbf{Does GraphThink improve long-horizon reasoning?}
Across baselines, performance drops markedly on long-horizon tasks, whereas GraphThink remains robust. Detailed analysis in Section~\ref{Varying_Horizons} further shows that, as task horizons increase, the baselines suffer pronounced degradation in planning success and reasoning fidelity, while GraphThink maintains stronger robustness and high inference efficiency. This highlights the benefit of the task graph for reliable and efficient long-horizon embodied planning.

\begin{table}[t!]
\centering
\caption{Qualitative comparison of long-horizon task planning performance with different training methods.}
\label{tab:case_long}
\scriptsize
\begin{tabular}{p{0.95\linewidth}}
\toprule
\rowcolor{instructionbg}
\textbf{Instruction:} Put a cleaned tomato slice in a bowl on the top shelf, and put a pan containing a heated potato on the dining table. \\
\midrule
\textbf{GraphThink (Ours):} 
(PickupObject, Knife), (SliceObject, Tomato), (PutObject, (Knife, SinkBasin)), (PickupObject, TomatoSliced), (CleanObject, (TomatoSliced, SinkBasin)), (PutPickObject, (TomatoSliced, Bowl)), (PutObject, (Bowl, Shelf)), (PickupObject, Potato), (HeatObject, (Potato, Microwave)), (PutPickObject, (Potato, Pan)), (PutObject, (Pan, DiningTable))\\
\textcolor{green!70!black}{Success: all goals are satisfied with valid transitions.}  \\
\midrule
\textbf{GRPO without Task Graph Rewards:} 
(PickupObject, Knife), (PickupObject, Potato), (HeatObject, (Potato, Microwave)), (PutObject, (Potato, Pan)), (PickupObject, Knife), (PickupObject, Tomato), (SliceObject, Tomato), (PutObject, (Knife, SinkBasin)), (PickupObject, TomatoSliced), (CleanObject, (TomatoSliced, SinkBasin)), (PutObject, (TomatoSliced, Bowl)), (PutObject, (Bowl, Shelf)), (PutObject, (Pan, DiningTable))

\textcolor{blue!80!black}{Error: invalid consecutive pickup/put transitions and missing pan pickup before table placement.} \\
\midrule
\textbf{SFT:} 
(PickupObject, Knife), (SliceObject, Tomato), (PutObject, (TomatoSliced, Bowl)), (PutObject, (Bowl, Shelf)), (PickupObject, Potato), (HeatObject, (Potato, Microwave)), (PutPickObject, (Potato, Pan)), (PutObject, (Pan, DiningTable))

\textcolor{blue!80!black}{Error: missing tomato cleaning and knife placement, with an invalid consecutive put transition.}  \\
\bottomrule
\end{tabular}
\end{table}

\textbf{Does GraphThink outperform other strategies under the same backbone?}
With the Qwen2.5-7B-Instruct backbone, SFT is strong on validation but drops sharply to 20.57\% on long-horizon tasks. Due to high similarity between ALFRED's training and validation sets, SFT handles in-domain short-term planning adequately but struggles with complex long-horizon tasks where planning hallucinations and missing steps prevail. In contrast, GraphThink achieves 90.04\% on long-horizon tasks, demonstrating that task graph-based RL generalizes effectively to complex task compositions without requiring costly supervised fine-tuning on expert-annotated long-horizon trajectories. To further qualitatively analyze long-horizon reasoning behavior, Table~\ref{tab:case_long} compares GraphThink with two in-domain training baselines, including SFT and GRPO without task graph rewards. 
The cases show that GraphThink reduces planning hallucinations and missing steps in complex instructions, demonstrating the effectiveness of task graph constraints and graph-based reward signals. See Appendix D for more cases. We also compare with task‑graph search baselines (Appendix C), where stepwise expansion often accumulates myopic errors and leads to sub‑optimal paths in long‑horizon settings, making naive search markedly less effective than our integrated LLM‑based planning. \\
\subsubsection{Ablations of our high-level planner}

\begin{table}[t!]
  \centering
  \small
  \caption{Success rate contributions of individual components in the proposed planner.}
   %\vspace{-0.1cm}
  \label{tab:ablation_planner}
    \setlength{\tabcolsep}{10pt}
  \begin{tabular}{lccc}
    \toprule
    \textbf{Method} & \makecell[c]{\textbf{Valid}\\ \textbf{Seen}} & \makecell[c]{\textbf{Valid}\\ \textbf{Unseen}} & \makecell[c]{\textbf{Long}\\ \textbf{Horizon}} \\
    \midrule
    w/o Task-Graph Prompt   & 77.93 & 74.11 & 40.69 \\
    w/o GRPO       & 49.88 & 52.64 & 33.67 \\
    w/o Verification & 95.12 & 95.09 & 86.25 \\
    \midrule
    All            & 96.22 & 96.56 & 90.04 \\
    \bottomrule
  \end{tabular}
  %\vskip -0.1in
\end{table}

We ablate the three task-graph modules in the high-level planner to assess their contributions (Table~\ref{tab:ablation_planner}):
(i) \textbf{Prompt with Task Graph.} Removing the task graph from the prompt during inference significantly reduces planning accuracy, showing that the graph enhances the LLM's ability to generate grounded plans and reduces hallucinations. (ii) \textbf{Graph GRPO with Task Graph.} Replacing the GRPO-trained model with the base Qwen2.5-7B-Instruct causes significant performance degradation, highlighting the importance of our task graph-based reinforcement learning framework. 
%relying solely on task graph prompting and verification feedback still outperforms non-training inference enhancement methods.
(iii) \textbf{Verification with Task Graph.} 
Verification feedback has little effect on the validation set, since initial plans generally satisfy graph constraints. However, on long-horizon tasks where constraints are sometimes violated, error feedback improves performance by 3.79\%. The verification loop is also bounded to at most 3 iterations, making it a low-cost guardrail against residual planning hallucinations. Notably, even without refinement, our planner remains competitive and still outperforms baselines.\\
\subsubsection{Ablations of GRPO}

\begin{table}[t]
\centering
\small
\setlength{\tabcolsep}{9pt}
\caption{Ablation studies of GRPO components and the two-stage instruction reward design.}
\label{tab:ablation_grpo}
\begin{tabular}{lccc}
\toprule
\textbf{Setting} & 
\makecell[c]{\textbf{Valid}\\ \textbf{Seen}} & 
\makecell[c]{\textbf{Valid}\\ \textbf{Unseen}} & 
\makecell[c]{\textbf{Long}\\ \textbf{Horizon}} \\
\midrule
\multicolumn{4}{c}{\textbf{\textit{Ablations of GRPO Components}}} \\
\midrule
w/o \(R_{\text{node}} + R_{\text{edge}}\) & 93.05 & 89.33 & 44.63 \\
w/o \(R_{\text{node}}\) & 93.54 & 91.66 & 76.50 \\
w/o \(R_{\text{edge}}\) & 89.88 & 85.03 & 50.50 \\
w/o \(R_{\text{inst}}\) & 84.27 & 84.66 & 55.59 \\
w/o \(R_{\text{format}}\) & 91.71 & 89.82 & 70.63 \\
w/o Task-Graph Prompt & 85.00 & 83.44 & 59.46 \\
\midrule
\multicolumn{4}{c}{\textbf{\textit{Ablations of the two-stage \(R_{\text{inst}}\)}}} \\
\midrule
Critical-only \(R_{\text{inst}}\) & 89.76 & 90.67 & 72.85 \\
GT-only \(R_{\text{inst}}\) & 91.71 & 90.06 & 68.84 \\
\midrule
Full GRPO & \textbf{95.12} & \textbf{95.09} & \textbf{86.25} \\
\bottomrule
\end{tabular}
\end{table}

\begin{figure}[t]
\centering
    \includegraphics[width=\columnwidth]{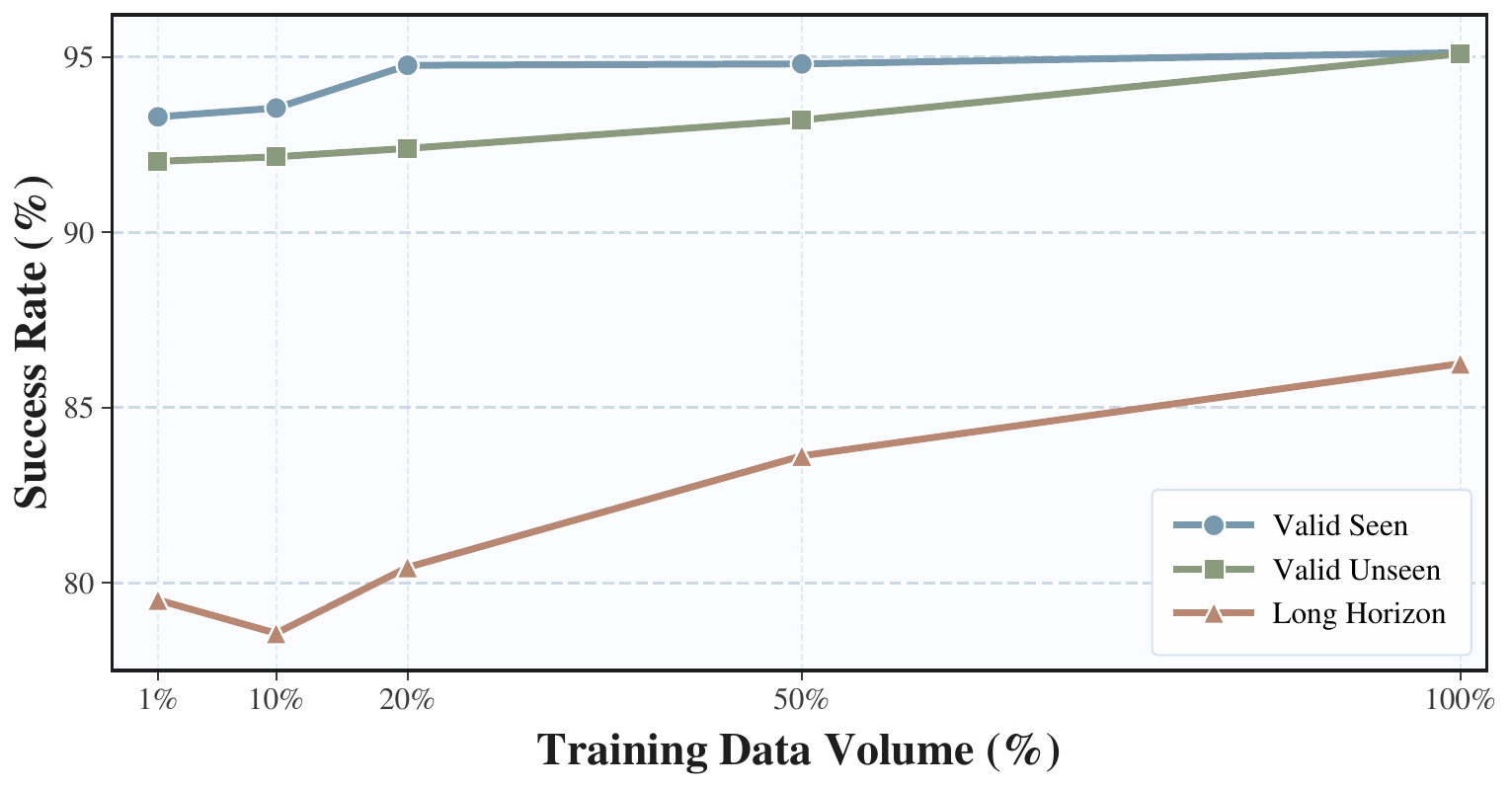}
    \caption{Ablation studies on training data volume.}
    \label{fig:training_data_volume}
\end{figure}

We conduct a quantitative ablation study to analyze key components in GRPO and further ablate the two-stage design of the instruction-following reward \(R_{\text{inst}}\), as shown in Table~\ref{tab:ablation_grpo}. The ``Full GRPO" setting indicates success rates with full rewards and the task graph prompt. Results here differ from Fig.~\ref{fig:plandata} as we remove validation feedback to isolate training effects. Our findings are:
(i) \textbf{Task graph reward enhances generalization and zero-shot learning:} Without the graph reward (i.e., ``w/o \(R_{\text{node}} + R_{\text{edge}}\)"), reliance on only the instruction reward restricts learning to task paths in the dataset. By contrast, the graph reward encourages exploration of all feasible subtask transitions, as the task graph encapsulates diverse possible paths. Its absence hinders long-horizon planning, confirming that the graph reward significantly improves generalization to \textcolor{black}{complex task combinations beyond training coverage}. Even without the instruction reward (i.e., ``w/o \(R_{\text{inst}}\)"), using only unlabeled data and graph rewards achieves an average 84.47\% on the valid set, demonstrating the graph reward's ability to guide high-level action learning and impart basic planning capabilities in zero-shot scenarios. (ii) \textbf{Edge-level transition constraints are especially important:} The fine-grained ablations of \(R_{\text{node}}\) and \(R_{\text{edge}}\) further show that removing either the node-level reward or the edge-level reward results in performance decline. Notably, the absence of \(R_{\text{edge}}\) causes a more significant drop, indicating that constraining subtask transitions through graph edges is crucial for grounded subtask decomposition. (iii) \textbf{Instruction reward aligns user intent:} Without the instruction following reward, the model may engage in reward hacking by solely generating plans that satisfy graph constraints while disregarding actual instruction requirements. These results underscore the essential role of the instruction following reward in ensuring semantic alignment between the generated plans and the original task instruction. Furthermore, using only critical-step coverage (i.e., ``Critical-only") or only exact match (i.e., ``GT-only") both underperform compared to the progressive design, especially on long-horizon tasks. This indicates that the two-stage design of \(R_{\text{inst}}\) improves semantic alignment without inducing overfitting to annotated paths. (iv) \textbf{Format reward supports structural output consistency:} When the format reward is ablated, a slight performance drop is observed. This suggests that although other rewards can partially guide the model to produce structured outputs, explicit format supervision helps ensure structure correctness. (v) \textbf{Incorporating the task graph into prompts enhances training effectiveness:} When we remove the explicit task graph from the prompt while keeping all other training and inference settings unchanged (i.e., ``w/o Task-Graph Prompt"), performance drops markedly on both the validation set and long-horizon tasks (average declines of 10.89\% and 26.79\%, respectively). These results confirm that the task graph not only aids model reasoning but also substantially improves training effectiveness.

\textbf{Impact of training data volume.}
To evaluate the impact of training data volume, we train the planning model using subsets corresponding to 1\%, 10\%, 20\%, and 50\% of the full training samples. These subsets cover all seven task types to ensure a fair representation of the training set. As shown in Fig.~\ref{fig:training_data_volume}, model performance improves with increasing amounts of training data. The performance gain is relatively modest on the validation set, indicating that our planning method can effectively strengthen short-task (\emph{i.e.}, validation-set) planning even with limited data. In contrast, the improvement is more pronounced on long-horizon tasks, suggesting that scaling the training data helps LLMs learn the richer action space required for complex planning.

\begin{figure*}[t!]
\vspace{-0.1cm}
    \centering
    \includegraphics[width=0.92\linewidth]{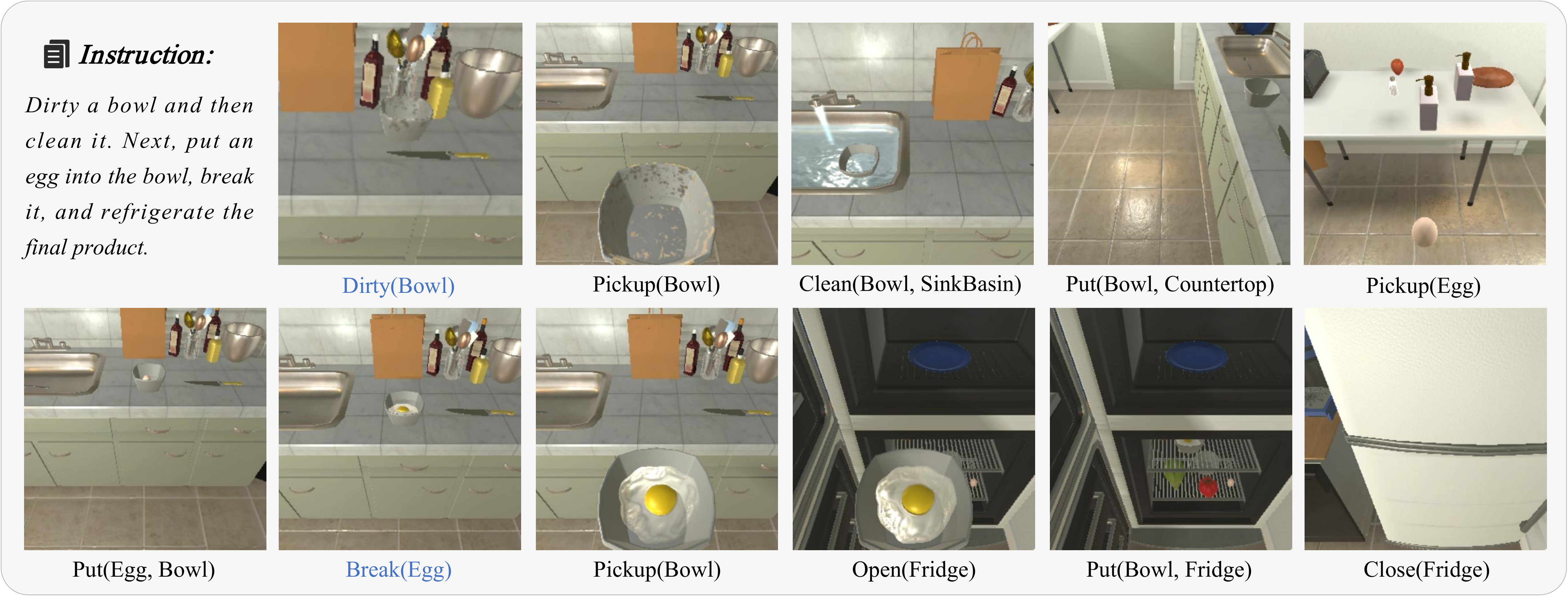}
    \caption{\textcolor{black}{An example of GraphThink executing a newly composed task involving unseen action primitives in AI2-Thor. The new tasks `DirtyObject' and `BreakObject' are highlighted in blue.}}
    \label{fig:newskill}
    \vspace{-0.1cm}
\end{figure*}

\subsection{Generalization to Novel Tasks and Environments}\label{novel_action_env}
To systematically evaluate GraphThink's generalizability beyond benchmark-specific settings, we conduct experiments along two complementary dimensions: (i) generalization to novel tasks in AI2-Thor, testing the ability to incorporate unseen action primitives; and (ii) cross-environment generalization to different task settings (\textit{i.e.},  WAH-NL~\cite{choi2024lota} and VirtualHome-HG~\cite{liu2025leap} in the VirtualHome environment), evaluating the transferability of our planning methodology to new domains. Following the best-performing baselines in Sec.~\ref{high-level-results}, we compare GraphThink with: (i) the leading RAG-enhanced LLM (GPT-5.2); and (ii) methods using the same backbone (Qwen2.5-7B-Instruct), including supervised fine-tuning (SFT) and few-shot prompting augmented with chain-of-thought (CoT) reasoning, to ensure a fair comparison.

\textbf{Generalization to novel tasks within AI2-Thor.}
We test GraphThink's ability to handle new action primitives by expanding its task graph with 9 new actions supported by the underlying AI2-Thor simulator but absent from ALFRED (e.g., BreakObject, FillObject, ThrowObject). Following a similar construction methodology as the Long-Horizon Dataset, we create 10 new task types combining these novel skills (e.g., Break\&Throw, Clean\&Fill\&Heat/Cool\&Place), with 400 samples. By extending the task graph with transition compatibility constraints, we evaluate GraphThink's zero-shot generalization to these novel tasks. As shown in Table~\ref{tab:generalization-AI2}, GraphThink substantially outperforms baselines on the newly composed tasks. Fig.~\ref{fig:newskill} visualizes an example execution of the ``Dirty\&Clean\&Break\&Cool\&Place'' task in the scene, illustrating how GraphThink composes newly introduced actions with existing manipulation skills.

\begin{table}[t!]
%\vspace{-0.1cm}
  \caption{\textcolor{black}{Generalization performance across new tasks.}}
  %\vspace{-0.1cm}
  \centering
  \setlength{\tabcolsep}{12pt}
  \small
  \begin{tabular}{l l c}
    \toprule
    \bf Method & \bf Model & \makecell[c]{\textbf{New} \textbf{Tasks}} \\
    \midrule
    RAG               & GPT-5.2                  & 65.75 \\
    Fs+CoT            & Qwen2.5-7B-Instruct      & 16.75 \\
    SFT               & Qwen2.5-7B-Instruct      & 16.25 \\
    \midrule
    GraphThink        & Qwen2.5-7B-Instruct      & \textbf{80.50} \\
    \bottomrule
  \end{tabular}
  \label{tab:generalization-AI2}
  %\vskip -0.2in
\end{table}

\begin{table}[t!]
%\vspace{-0.3cm}
  \caption{\textcolor{black}{Cross-Environment Generalization to VirtualHome.}}
  %\vspace{-0.1cm}
  \centering
  \setlength{\tabcolsep}{2.6pt}
  \small
  \begin{tabular}{l l c c}
    \toprule
    \bf Method & \bf Model & \bf{WAH-NL} & \textbf{VirtualHome-HG}\\
    \midrule
    RAG               & GPT-5.2            & 92.00 & 88.89 \\
    Fs+CoT  & Qwen2.5-7B-Instruct  & 32.00 &  17.78 \\
    SFT               & Qwen2.5-7B-Instruct      & 87.00 & 78.89 \\
    \midrule
    Ours(ALF)  & Qwen2.5-7B-Instruct            & 90.00 & 76.67 \\
    Ours(VH)   & Qwen2.5-7B-Instruct           & \textbf{93.00} & \textbf{88.89}\\
    \bottomrule
  \end{tabular}
  \label{tab:generalization-VH}
\end{table}

\begin{figure*}
    \centering
    \includegraphics[width=0.92\linewidth]{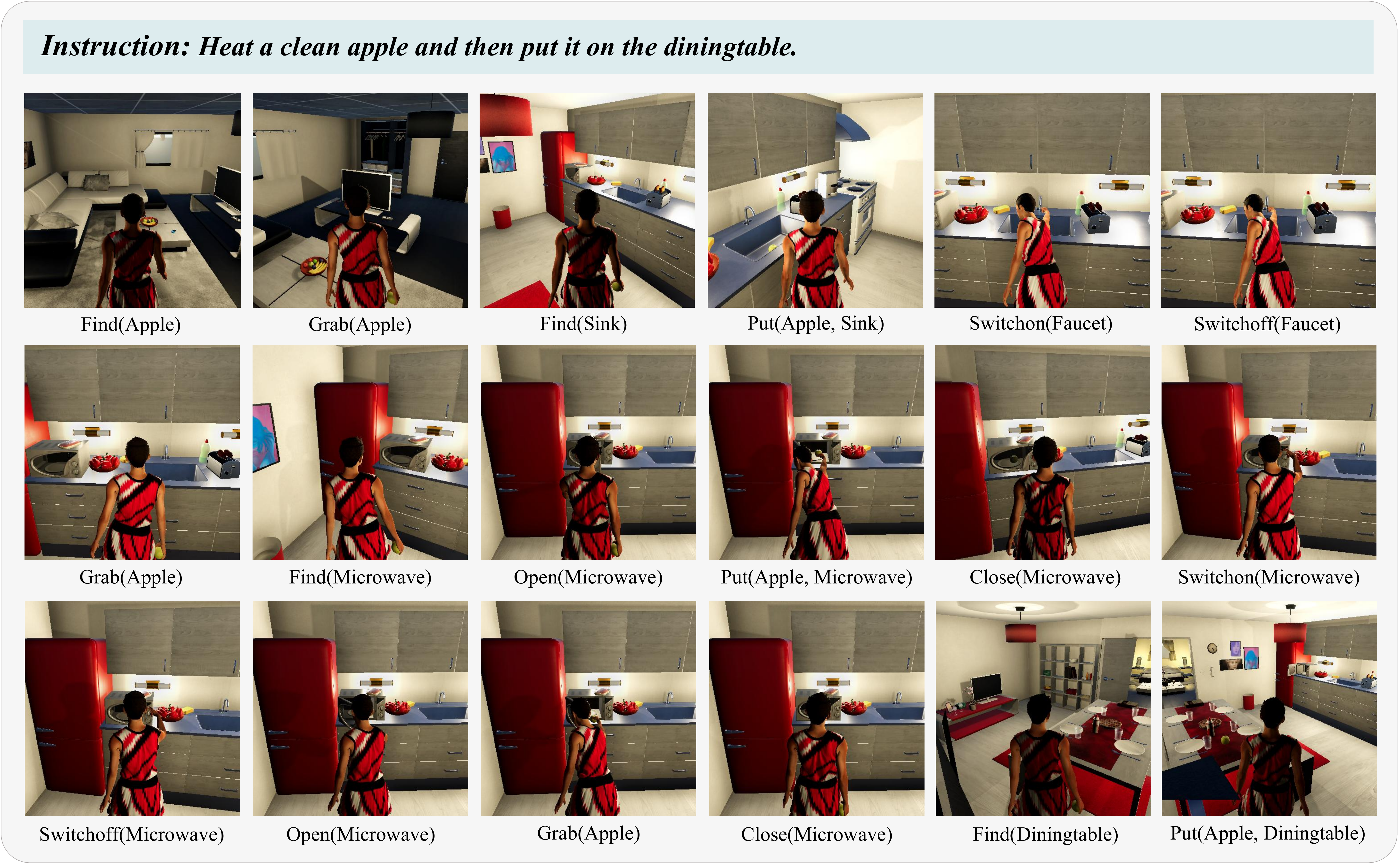}
    \caption{An example of GraphThink for the task `Heat a clean apple and then put it on the dining table' in VirtualHome.}
    \label{fig:vh}
\end{figure*}

\textbf{Cross-Environment Generalization to VirtualHome.} To further validate the versatility of the proposed GraphThink in broader robotic task applications, we test it on two additional benchmarks built on the VirtualHome simulator, including 100 rearrangement tasks from WAH‑NL and 90 VirtualHome‑HG tasks covering cooking, cleaning, and laundry. We test two adaptation strategies:
\textbf{(i) Domain-specific adaptation:} VirtualHome has a partially different action space and interaction primitives from ALFRED. In this case, we construct a new task graph from VirtualHome's action primitives, and the planner is trained with our framework using in-domain data.
\textbf{(ii) Cross-domain transfer:} Although the low-level primitives differ, most high-level subgoal semantics can be shared across embodied environments. We apply the ALFRED-trained planner and map abstract subtasks to VirtualHome's action programs via a predefined translation layer that aligns semantic action types between domains. If novel subtask types that do not exist are encountered (e.g., laundry), the task graph can be extended with corresponding nodes to support the new skills.

Baselines are fairly compared using corresponding in-domain VirtualHome data. Results in Table~\ref{tab:generalization-VH} show that GraphThink with domain-specific adaptation (VH) delivers the best performance, while the cross-domain variant (ALF) remains competitive, indicating that our high-level planner generalizes well to VirtualHome despite differences in action space and task semantics. These results further suggest that (i) high-level planning priors learned in the ALFRED domain are reusable across embodied simulators, and (ii) re-instantiating the planning framework for the target domain further enhances cross-environment generalization, validating the effective cross-domain transferability of GraphThink. Fig.~\ref{fig:vh} further provides a qualitative execution example in the VirtualHome environment.\\

\subsection{Performance Analysis across Varying Task Horizons}\label{Varying_Horizons}
\begin{figure*}[!ht]
    \centering
    \includegraphics[width=1\linewidth]{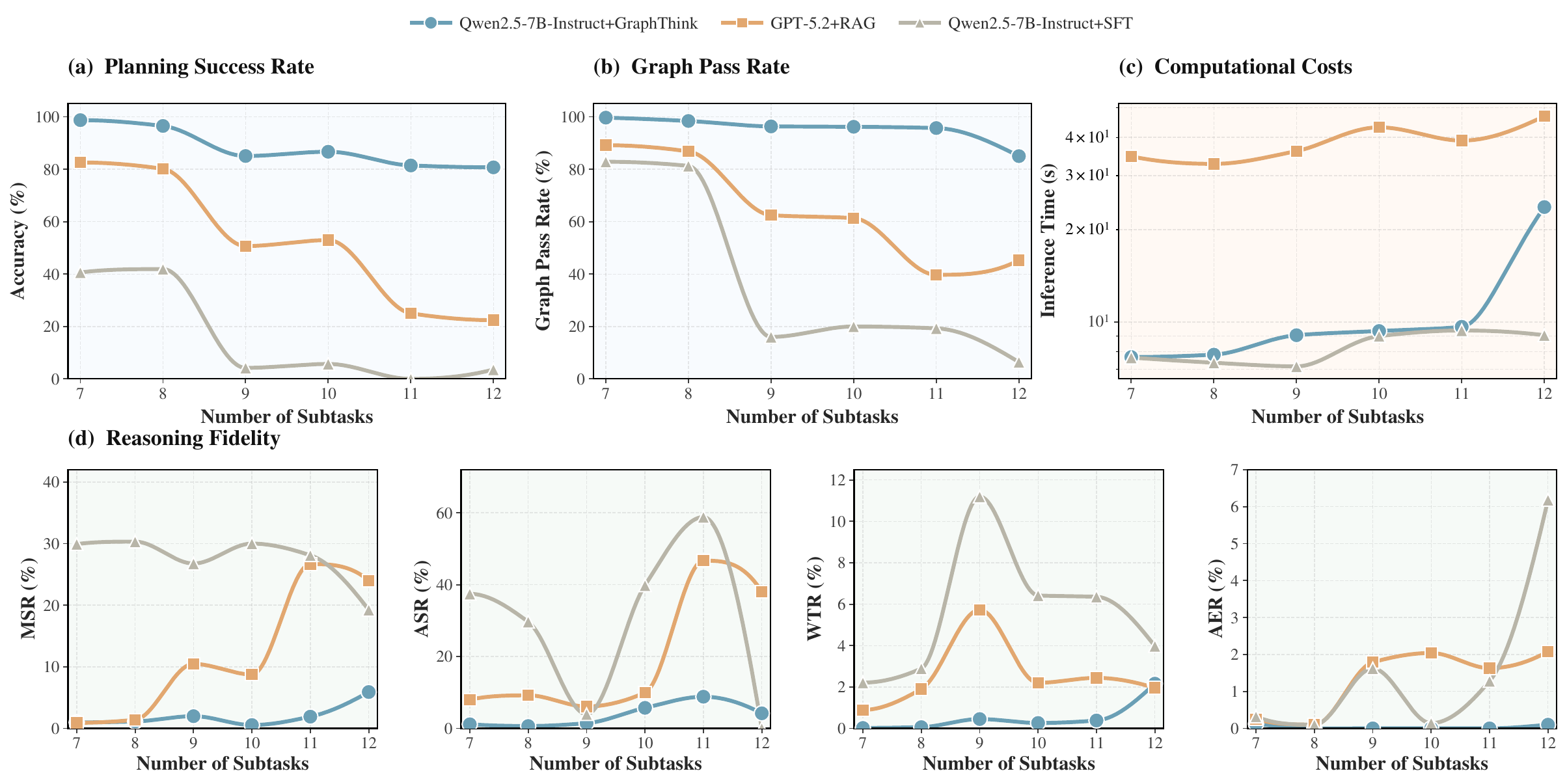}
    \caption{
    Performance comparison across increasing task horizons.
    Subfigure (a) reports planning success rate by accuracy, and subfigure (b) shows graph pass rate.
    Subfigure (c) compares computational costs measured by inference time.
    Subfigure (d) presents fine-grained reasoning-fidelity metrics, including missing step rate (MSR), additional step rate (ASR), wrong transfer rate (WTR), and affordance error rate (AER).
    Qwen2.5-7B-Instruct+GraphThink shows stronger robustness and planning fidelity than the baselines, while achieving substantially lower inference time than GPT-5.2+RAG on long-horizon tasks.
    }
    \label{fig:horizon_scaling}
\end{figure*}

To comprehensively evaluate the robustness and scalability of our approach, we analyze GraphThink's performance across tasks of increasing complexity (7–12 subtasks), comparing it against two strong baselines: GPT-5.2 enhanced with RAG and a SFT model based on the same Qwen2.5-7B-Instruct backbone.
The results are summarized in Fig.~\ref{fig:horizon_scaling}. All metrics are reported as averages. We use the following abbreviations: Acc (Accuracy), GPR (Graph Pass Rate), MSR (Missing Step Rate), ASR (Additional Step Rate), WTR (Wrong Transfer Rate), and AER (Affordance Error Rate). MSR, ASR, WTR, and AER are computed per task instance and then averaged over all tasks with the same horizon length, reflecting per-task reasoning fidelity.

\textcolor{black}{
(i) \textbf{Performance Scaling}: As shown in Fig.~\ref{fig:horizon_scaling}(a), Qwen2.5-7B-Instruct+GraphThink shows only moderate degradation as task horizons increase, with accuracy decreasing from 98.73\% to 80.71\%. Meanwhile, Fig.~\ref{fig:horizon_scaling}(b) shows that its graph pass rate remains at or above 85\% across all horizons. This contrasts sharply with the baselines: GPT-5.2 with RAG drops from 82.59\% to 22.40\%, while Qwen2.5-7B-Instruct+SFT drops from 40.63\% and remains near zero from 9 subtasks onward.}

\textcolor{black}{
(ii) \textbf{Computational Costs}: As shown in Fig.~\ref{fig:horizon_scaling}(c), Qwen2.5-7B-Instruct+GraphThink maintains stable inference time ($\sim$7--10s) for tasks with 7--11 subtasks, demonstrating well-controlled overhead. Even at 12 subtasks, the average inference time only increases to 23.7s, which remains lower than GPT-5.2 with RAG (32.78--46.90s). Although Qwen2.5-7B-Instruct+SFT has low inference time, its success rate and reasoning fidelity degrade severely on long-horizon tasks. These results collectively validate the robustness, efficiency, and reasoning quality of Qwen2.5-7B-Instruct+GraphThink on increasingly complex long-horizon tasks.}

\textcolor{black}{
(iii) \textbf{Reasoning Fidelity}: Fig.~\ref{fig:horizon_scaling}(d) reports four fine-grained error metrics following~\cite{li2024embodied}, which help identify specific weaknesses in LLM planning. Qwen2.5-7B-Instruct+GraphThink consistently achieves superior planning quality, with remarkably low wrong transfer rates (0.02--2.16\%) and near-zero affordance errors, indicating accurate action--object grounding under graph constraints.
Although Qwen2.5-7B-Instruct+GraphThink shows moderate fluctuations in missing and additional step rates on longer horizons, these errors remain well controlled compared with the baselines. For example, its ASR peaks at 8.81\% at 11 subtasks, whereas GPT-5.2 with RAG reaches a much higher ASR of 46.67\% at the same horizon.}

% novel tasks 

% novel environments

\section{Conclusion}

We introduce GraphThink, a general graph-enhanced planning framework that improves LLM-based embodied task planning. Our method reduces planning hallucinations by grounding reasoning in the structured task graph and ensuring plan executability through event-driven replanning with dynamic scene graphs. Experimental results on ALFRED validate the effectiveness of our components, while tests on a new long-horizon dataset demonstrate strong reasoning and generalization capabilities. Additional experiments show promising zero-shot learning potential. \textcolor{black}{Future work could extend this approach to asynchronous planning or multi-robot collaborative systems through temporal-state edge augmentation or multi-layer graph construction.} 

\bibliographystyle{IEEEtran} 
\bibliography{example_paper}

%%%%%%%%%%%%%%%%%%%%%%%%%%%%%%%%%%%%%%%%%%%%%%%%%%%%%%%%%%%%%%%%%%%%%%%%%%%%%%%
%%%%%%%%%%%%%%%%%%%%%%%%%%%%%%%%%%%%%%%%%%%%%%%%%%%%%%%%%%%%%%%%%%%%%%%%%%%%%%%

%{\appendices
%\section*{Proof of the First Zonklar Equation}
%Appendix one text goes here.
% You can choose not to have a title for an appendix if you want by leaving the argument blank
%\section*{Proof of the Second Zonklar Equation}
%Appendix two text goes here.}

\vfill

\end{document}

%% file: math_commands.tex
\usepackage{amsmath,amsfonts,bm}

\def\eqref#1{equation~\ref{#1}}
\def\1{\bm{1}}

\DeclareMathAlphabet{\mathsfit}{\encodingdefault}{\sfdefault}{m}{sl}
\SetMathAlphabet{\mathsfit}{bold}{\encodingdefault}{\sfdefault}{bx}{n}